\documentclass[journal]{IEEEtran}
\usepackage{lineno,hyperref}
\usepackage{bbm}
\usepackage{bm}
\usepackage{amsfonts}
\usepackage{amsmath}
\usepackage{mathrsfs}
\usepackage{amssymb}
\usepackage{enumerate}
\usepackage{graphicx}
\usepackage{subfigure}
\usepackage{booktabs}
\usepackage{graphicx}
\usepackage{setspace}
\usepackage{algorithm}
\usepackage{algorithmic}
\usepackage{setspace}
\usepackage{multirow}
\usepackage{color}
\usepackage{cite}
\usepackage{array}
\usepackage{times}
\usepackage{mathptmx}
\usepackage{dsfont}
\ifCLASSINFOpdf
\else
\fi
\begin{document}
\title{Person Text-Image Matching via Text-Feature Interpretability Embedding and External Attack Node Implantation}
\author{Fan Li, Hang Zhou, Huafeng~Li, Yafei Zhang, and Zhengtao Yu \IEEEmembership{}
\thanks{This research was supported by the National Natural Science Foundation of China (Nos. 62276120, 61966021, 61562053), the Basic Research Project of Yunnan Province(No. 202101AT070136) (\emph{Corresponding authors: Huafeng Li}.)}
\thanks{F. Li, H. Zhou, H.F. Li, Y.F. Zhang and Z.T. Yu affiliated with the Faculty of Information Engineering and Automation, Kunming University of Science and Technology, Kunming 650500, China.(e-mail: lifan198686@163.com (F. Li); zhhylysys@yeah.net (H. Zhou); lhfchina99@kust.edu.cn (H. Li); zyfeimail@163.com (Y. Zhang); ztyu@hotmail.com (Z. Yu))}
}
\markboth{Journal of \LaTeX\ Class Files}%
{Shell \MakeLowercase{\textit{et al.}}}
\maketitle
\begin{abstract}
Person  text-image matching, also known as text-based person search, aims to retrieve images of specific pedestrians using text descriptions. Although person text-image matching has made great research progress, existing methods still face two challenges. First, the lack of interpretability of text features makes it challenging to effectively align them with their corresponding image features. Second, the same pedestrian image often corresponds to multiple different text descriptions, and a single text description can correspond to multiple different images of the same identity. The diversity of text descriptions and images makes it difficult for a network to extract robust features that match the two modalities. To address these problems, we propose a person text-image matching method by embedding text-feature interpretability and an external attack node. Specifically, we improve the interpretability of text features by providing them with consistent semantic information with image features to achieve the alignment of text and describe image region features. To address the challenges posed by the diversity of text and the corresponding person images, we treat the variation caused by diversity to features as caused by perturbation information and propose a novel adversarial attack and defense method to solve it. In the model design, graph convolution is used as the basic framework for feature representation and the adversarial attacks caused by text and image diversity on feature extraction is simulated by implanting an additional attack node in the graph convolution layer to improve the robustness of the model against text and image diversity. Extensive experiments demonstrate the effectiveness and  superiority of text-pedestrian image matching over existing methods. The source code of the method is published at \textcolor{blue}{\url{https://github.com/lhf12278/SAA}}.
\end{abstract}
\IEEEpeerreviewmaketitle
\section{Introduction}
Although video surveillance equipment can be used in all corners of a city, its application can be hindered by the special characteristics of certain occasions. In such scenarios, it is likely that only eyewitnesses are available, resulting in the inability to use photos of suspects to search for their whereabouts. Therefore, person text-image matching has been proposed to address this problem[1]. This technique retrieves the target pedestrian image using the text description of the pedestrian's appearance provided by the eyewitness. This is a typical cross-modal image retrieval problem. Compared with person re-identification[50,47,45,48,52,49], it does not require the target pedestrian image as the query image and can avoid the difficulties caused by the lack of surveillance devices for pedestrian search and thus has wide application potential.

\begin{figure}[t!]
\centering
\subfigure{\includegraphics[height=1.7in, width=3in]{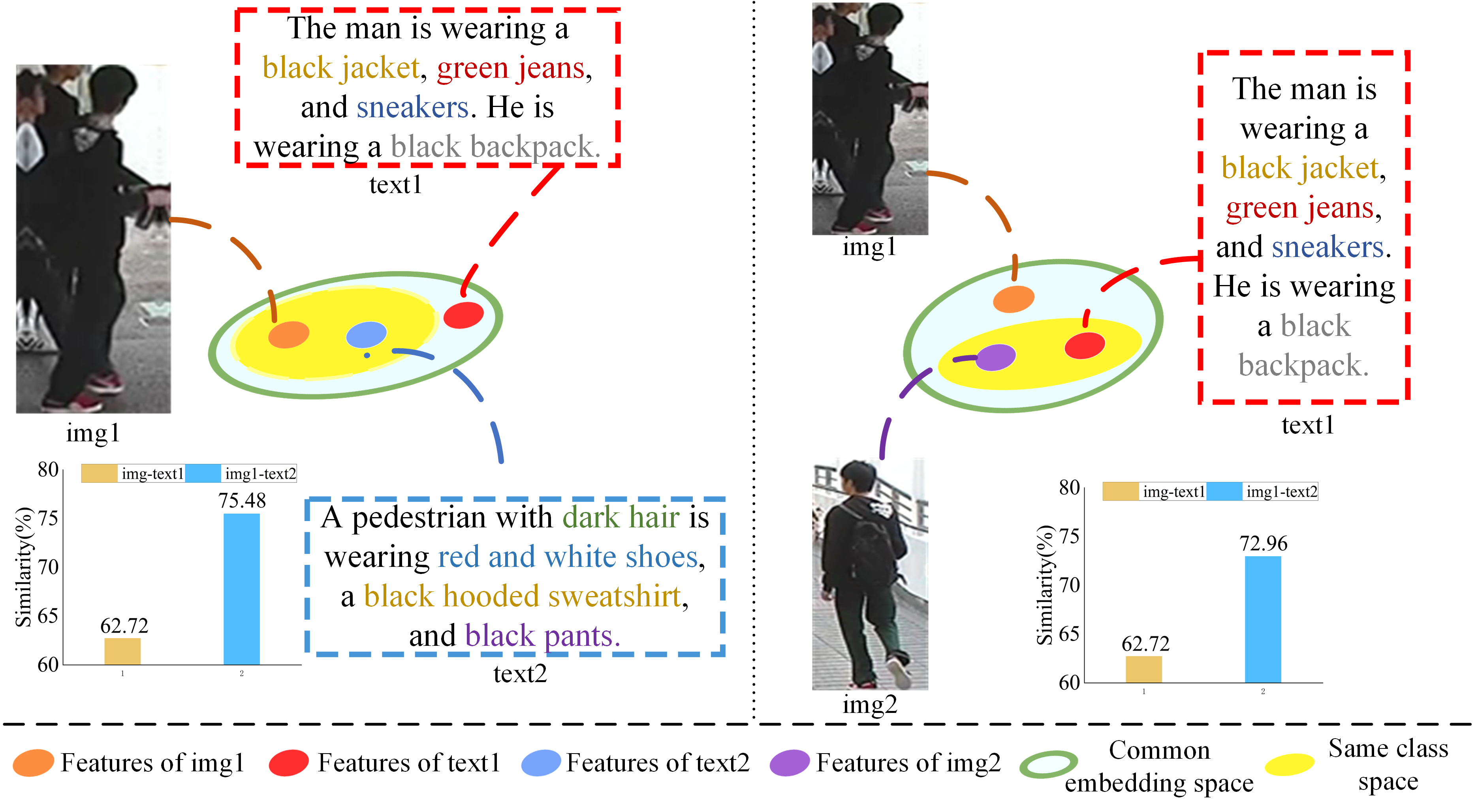}}
\caption{Illustration of the influence of person image and text diversity on text-image matching. Due to the diversity of text and person images, text1-img1, text2-img1, text1-img2 and text2-img2 show different similarities. This indicates that the diversity of the person text-image contributes a significant amount of uncertainty to person text-image matching.}
\label{png1}
\end{figure}
In person text-image matching, the alignment of keywords in a sentence with the described object in images is one of the key factors affecting recognition performance. To address this issue, numerous effective feature-alignment methods have been proposed in recent years. These methods can be broadly classified into those based on similarity relation metrics \cite{2,13}, external knowledge assistance \cite{5,8,12,15}, local relation correspondence \cite{9,10,6,3,14} multi-headed attention  \cite{7,11} etc.  Similarity relationship metric-based methods measure the similarity between noun phrases and local patches of images, and determine their relationships based on the predicted weights. In external knowledge-assisted-based approaches, the semantics of the human body \cite{8}, pedestrian pose \cite{5}, and pedestrian attributes  \cite{12,15} are often used as auxiliary information in the alignment of text and visual features. Methods based on local relational correspondence often achieve the local alignment of text and person image features through a specific function relationship or attention mechanism \cite{13}. In contrast to the attention mechanism used in methods based on local relational correspondence, methods based on multi-head attention usually assign different semantics to each head to align it with a specific region of the image.

Methods based on similarity relationship metrics tend to introduce background noise, which affects feature quality. Although external knowledge-assisted approaches can alleviate this problem, the acquisition of external knowledge, such as the semantics of the human body and pedestrian pose, requires the use of a human semantic parsing model and pedestrian keypoint detection model. However, achieving human semantic parsing and keypoint detection remains an open problem. Because of the diversity of pedestrians in image spatial locations, feature alignment methods based on local relational correspondence are subject to the problem that an image patch may correspond to multiple keywords or noun phrases and that a noun phrase or keyword is also associated with multiple patches. This poses a challenge to the establishment of a local correspondence between the text and image. Although attention-based methods \cite{3,6,13} can effectively alleviate this problem, they require high computational effort and low implementation efficiency.

In practice, a one-person image often corresponds to several different textual descriptions. These texts may show different structures and noun phrases,  as well as different images of the same identity, as shown in Fig. \ref{png1}. Such differences bring challenges to the matching of text and pedestrian images. For example, for the two texts, that is, text1 and text2 in Fig. \ref{png1}, the similarity between image1 and text1 is approximately 62\%, whereas the similarity between image1 and text2 has reached 75\%. This suggests that differences in sentence structure are one of the factors hindering the matching of text pedestrian images. In addition, as shown in the right half of Fig. \ref{png1}, the similarity between text1 and image2 is 72\%; whereas, the  similarity between text1 and image1 is only 62\%. This indicates that the diversity of pedestrian images with the same identity also leads to greater instability in cross-modal matching. However, this issue has not received enough attention from the community. Although some text-image matching methods have focused on this problem \cite{16,17}, it has not been effectively resolved.

In this study,  we propose a novel text-feature interpretability embedding and external attack node implantation method for person text image matching. This method fully considers the relationship between key information in sentences and extracts features corresponding to local regions of pedestrian images from the text features extracted using BERT \cite{30}.  In this process, by giving the text features the same semantics as a local area of the pedestrian image, embedding the interpretability of text features and fine-grained alignment of the text and its corresponding pedestrian image are achieved. In feature construction for text and person image matching, the proposed method is different from the existing method \cite{14}, which obtains global features by pooling aligned local features. This causes misalignments between text and visual features. The proposed method achieves semantic consistency and interpretability of global features by placing features with the same semantics at the same spatial location to construct the final features for person text-image matching.

Considering the impact of sentence structure differences as well as pedestrian image diversity on person text-image matching, this paper introduces the idea of adversarial attack and defense in this field for the first time and proposes a graph convolution method for attack node implantation. In this method, we view the sentence structure and person image diversity as caused by perturbations that lead to the degradation of model performance. If the feature extraction network can defend against such perturbations, then the problems caused by the sentence structure as well as pedestrian image diversity can be effectively mitigated. To this end, we propose the introduction of a learnable external node in the structural framework of graph convolution and use it as an attack node to disturb the features of each node. To ensure that the node has the ability to attack, we propose using the gradient ascent method. In the graph convolution process, an attack node is used to inject the perturbation into each node. After obtaining the perturbation-injected attack node, we used it to attack the person text-image matching model, thereby making it capable of defending against the attack from perturbation. Thus, the proposed method is endowed with strong defense against perturbations caused by text and pedestrian image diversity. Experiments show that the method proposed can effectively align the local features of text and pedestrian images, alleviate the challenges caused by the diversity of text and pedestrian images, and improve the cross-modal matching performance.

The contributions of this study and the advantages of the proposed approach are as follows.
\begin{itemize}
\item We propose a person text-image feature semantic alignment method from the viewpoint of feature interpretability. In this method, we use the local features of a pedestrian image to guide the extraction of local text features to ensure that they have the same semantics. Based on these local features endowed with specific semantics, global features describing pedestrian text and images are constructed in a specific semantic order, thereby achieving semantic alignment of the global features of text and the image.

\item
To address the challenges posed by sentence and pedestrian image diversity to text image matching, this paper proposes to view the diversity of text and corresponding pedestrian images as caused by adversarial perturbation. The idea of adversarial attacks and defense is proposed for the first time to solve this problem. Technically, we inject a perturbation on the node by introducing an external node into the graph convolution and generating an attack node. With the generated attack node, we empower the matching model to defend against perturbation through adversarial training.

\item
The proposed method considers the diversity of both textual and pedestrian images. Therefore, it can not only alleviate the problems caused by text diversity but also improve the robustness of the model to pedestrian poses, camera views, and style changes. The method presented in this paper shows excellent performance on different datasets, which proves its effectiveness.

\end{itemize}

The remainder of this paper is organized as follows. Section~\ref{sec:related} reviews related works, and Section~\ref{sec:method} introduces the proposed method in detail. The experimental results are presented and analyzed in Section~\ref{sec:experiment}, and Section~\ref{sec:conclusion} concludes the paper.
\section{Related Work}
\label{sec:related}
\subsection{Text-based Image Retrieval}
Text-based image retrieval is also known as text-image matching \cite{46}. This refers to using a text description to retrieve images that are consistent with the text description \cite{1}. In this task, solving the modal gap between text and image features and achieving the alignment of semantic information are key factors affecting the performance of the model. To this end, researchers have proposed a series of effective methods that can be divided into modality-interaction-based methods \cite{13} and modality-independent methods \cite{14, 46} according to the testing process. Modal-interaction-based matching methods often obtain multimodal feature representations by performing an intermodal crossover of text-image features. Such methods usually achieve a better matching performance with the aid of complex cross-attention. However, such methods require a combination of text and all images to be matched to form text-image pairs. Each text-image pair must undergo feature extraction and multimodal feature crossover, which increases the computational burden and is not conducive to the deployment and application of the model in practical scenarios. The modality-independent method uses  two different models to extract features from text and images. In this process, the features of the two modalities do not interact or merge and mainly rely on the constraints of the loss function to achieve cross-modal matching. In practical applications, the text and image to be matched require only one feature extraction; therefore, it can adapt to large-scale text image retrieval tasks. However, these methods do not consider the impact of image and text diversity on the matching performance.

To solve the problem of sentence structure diversity, Song et al. \cite{18} proposed a polysemous instance embedding network. It maps text and images to $K$ spaces through multihead attention. In these $K$ spaces, the relational correspondence between multiple feature semantics of the text and image is realized to solve the problem of instance ambiguity. Because  multiheaded self-attention is involved in this method, it cannot eliminate  the time consumption caused by attention computation. Chun et al. \cite{17} proposed a probabilistic cross-modal embedding method for text-based image retrievals. This method represents samples from different modalities as probability distributions to represent several relationships in a joint embedding space. Although this method can solve the problem of sentence and image diversity, due to the large difference between text-based image retrieval and person text-image matching, directly introducing such methods into person text-image matching may lose their original performance. In contrast to the above methods, this study considers the diversity of textual and visual features caused by the perturbation of features. By improving the robustness of the model to perturbation, the model is able to defend against the changes in features caused by  text and image diversity.

\subsection{Person Text-Image Matching}
Although person text-image matching belongs to the category of text-based image retrieval, significant differences still exist between them. An image in text-based image search often contains multiple objects, and there is a certain interrelated relationship between these objects. In text-person image matching, a person image often contains only one pedestrian. To obtain accurate matching results, the model must be able to extract fine-grained and modality-independent features from pedestrian images and text. Therefore, person text-image matching is more challenging than text-based image retrieval. For person text-image matching, Li et al. \cite{1} proposed recurrent neural network with gated neural attention mechanism (GNA-RNN) to perform text-person image search \cite{1}. Moreover, a large-scale person description dataset named the CUHK person description dataset (CUHK-PEDES) was constructed, which contains detailed natural language annotations and person samples from multiple datasets. Subsequently, Li et al. \cite{16} proposed an identity aware two-stage network, where stage-1 can efficiently screen simple incorrect matchings, while stage-2 refines the matching results by using co-attention to align local regions of words and images.

To address the cross-modal matching of text and pedestrian images, similarity relation metrics (SRM) based methods \cite{2,13,16}, external knowledge assistance (EKA)-based methods \cite{8,5,12,15}, and multi-granularity relational correspondence (MRC)-based feature alignment methods \cite{9, 10,6, 3, 14,7} are presented. The SRM-based method constructs the relationship between noun phrases and image blocks by measuring their similarities. In this process, it is necessary to measure the similarity between the query text and all the images in the gallery. This requires the text and image to be matched to re-extract features when calculating the similarity, which significantly reduces the test efficiency. The EKA-based method requires external information to assist in improving the model performance. Common external information in person text-image matching includes pedestrian key points, pedestrian attributes, and body semantic segmentation results. However, to obtain this external information, an additional specific model must be used, which results in a high degree of dependence on the model. The MRC-based method often constructs feature representations for person text image matching based on aligned multigranularity features. This process can easily lead to confusion in fine-grained semantic information, which is detrimental to the matching task. In contrast to the above methods, to achieve semantic consistency features, this study proposes mapping the global features of text into the feature space corresponding to a local pedestrian image through nonlinear mapping. Simultaneously, a  graph with a consistent structure is constructed such that the nodes from the text and pedestrian image at the same position in the graph have consistent semantic information. Based on the aligned node features, we constructed a global feature representation with consistent semantic information to match text and pedestrian images.
\begin{figure*}[t!]
\centering
\includegraphics[width=6.4in,height=3.5in]{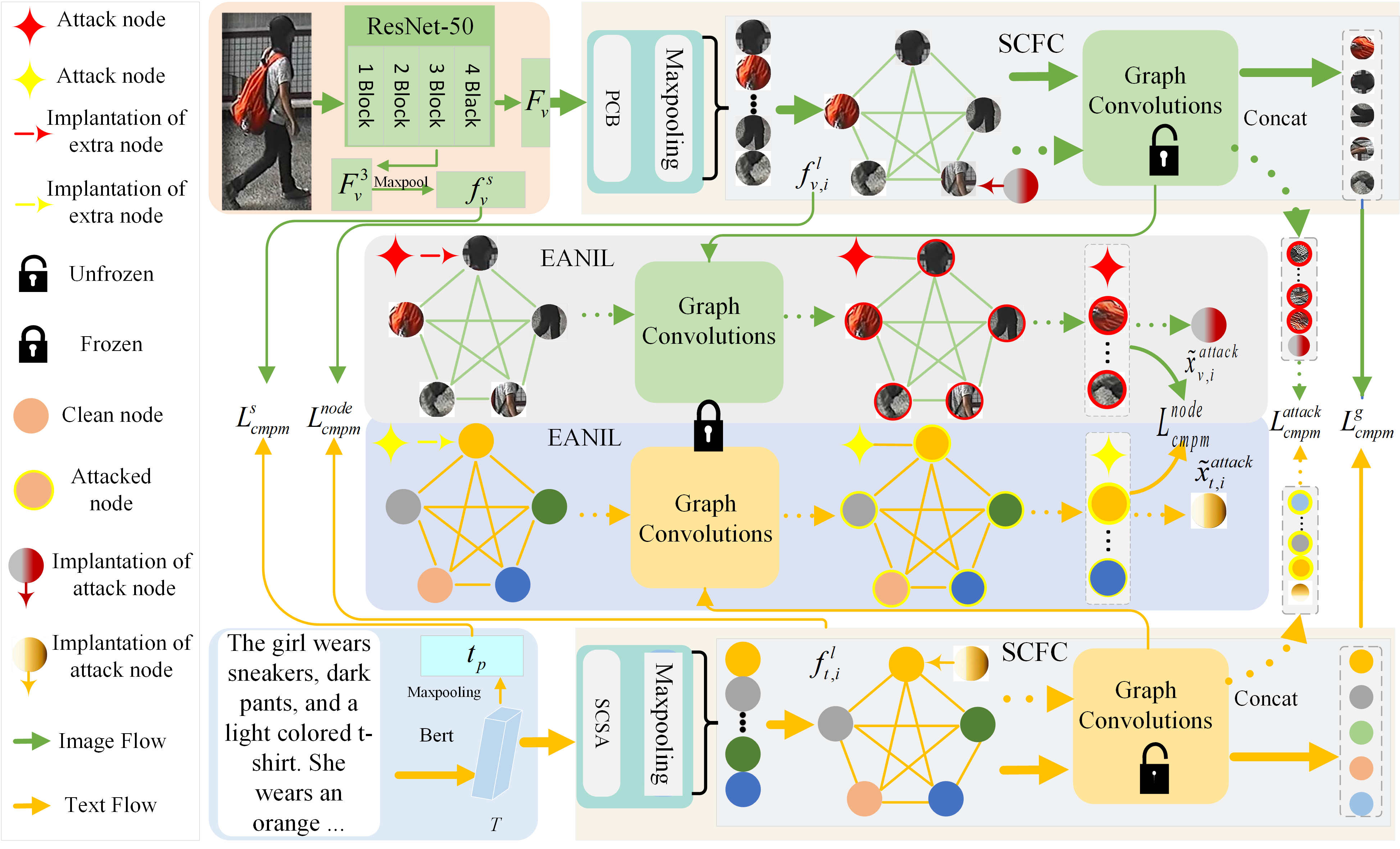}
\caption{Overview of the proposed method. The input text and person image are sent respectively to BERT and ResNet50 to obtain the initial global feature $\bm T$ and feature graph $\bm F_v$. $\bm T$ is sent to SCFC to ensure that the results have the same semantic information and graph structure as the patch features of $\bm f_{v,i}^l$. In the implanting and learning of external attack node, the attack node should be able to degrade the performance of the model after graph convolution. With the learned attack node, the robustness of matching models to text and image diversity is improved in adversarial training.}
\label{png2}
\end{figure*}
\subsection{Adversarial Attack and Defense}
Adversarial attacks invalidate a deep learning model by adding a small-magnitude perturbation to the input samples. This concept was first proposed by Szegedy et al. \cite{41}. With the increasing attention paid by researchers to the explainability of deep learning, adversarial attacks and defense have become research hotspots in recent years \cite{20,21,51}. In graph neural network, an attacker can perform adversarial attacks on a graph by perturbing the nodes, edges, and properties of nodes \cite{20,21,22,23}. In the real world, modifying the edges or attributes of an existing graph is difficult because of the lack of access to the database that stores graph data [32]. Node injection attacks can achieve adversarial attacks by injecting malicious nodes without modifying the existing graph \cite{24,25,26,27}. In particular, the node injection poisoning attack (NIPA) proposed by Sun et al. \cite{24} uses a hierarchical Q-learning network to sequentially generate edges and labels of malicious nodes. However, this method cannot generate the attributes of the injected nodes, which are extremely important for evading attacks from malicious nodes, thus limiting further performance improvements. Wang et al.  \cite{26} proposed an approximate closed-form attack solution for graph convolutional networks (GCNs) using the approximate fast gradient sign method (AFGSM), this method cannot be applied to all GCNs. Because both NIPA and AFGSM are proposed in poison settings, a defense model needs to be retrained for each attack. Unlike such methods, Zhou et al. \cite{27} proposed a topological defective graph injection attachment (TDGIA) method. This method can attack large-scale graphs that the NIPA cannot handle. Chen et al. \cite{28} proved that a node injection attack is more harmful than a graph modification attack owing to its high flexibility.

To address the challenge of cross-modal matching caused by the diversity of text and pedestrian images, we proposed an adversarial attack and defense method inspired by recent advances in adversarial attacks on graphs \cite{24,25,26,27}. This approach treats the diversity of sentences and images as a perturbation that causes changes in the features of text and its corresponding image, and designs an adversarial attack on a graph to simulate this perturbation. Owing to  this design, the robustness of the matching model to the diversity of text and the corresponding image was improved. To the best of our knowledge, this is the first approach to introduce the concept of adversarial attack and defense to person text image matching. This method can realize adversarial training of a model without generating adversarial samples; therefore, it is more practical.
\section{Proposed Method}
\label{sec:method}
\subsection{Overview}
The proposed approach views the diversities of text and image as adversarial attack information from the perturbation of features and improves the robustness of the model by making it more defensible against perturbations caused by diversity. As shown in Fig. \ref{png2}, the proposed framework consists of two main parts: semantic consistency feature construction (SCFC) and external attack node implantation and learning (EANIL). SCFC is used to solve the problem of misalignment between text and image features. The proposed method constructs semantic-aligned global features based on nodes with consistent semantics and graph structures, which facilitates subsequent person text-image matching. EANIL is primarily used to learn adversarial attack nodes that can degrade the model's performance. After the adversarial attack node is trained, adversarial training is employed to further train the model to improve the robustness of person-text image matching.

\subsection{Semantic Consistency Feature Construction}
Because the semantic information of each region of the pedestrian image is clear, in the proposed method, we use pedestrian image local features to guide the learning of text features, so that the text has the same semantics and structure as the corresponding image features, ensuring the interpretability of text features. In the extraction of person image features, ResNet50 is used as the backbone and is denoted as $\bm E_{bv}$. As shown in Fig. \ref{png2}, we represent the features extracted from the first three convolutional  layers of ResNet50 as  $\bm F_v^3 \in {\mathbb{R}^{H \times W \times C}}$, where $H$, $W$ and $C$ denote the length, width, and number of channels in the feature map, respectively. The feature output from the last convolutional layer is denoted as $\bm F_v \in {\mathbb{R}^{H \times W \times C}}$ and is sliced horizontally into $N$ patches. The feature maps of the  $l$-th patch of the $i$-th image in a batch are represented as  $\bm P_{v,i}^l \in {R^{H/N \times W \times C}}$, where $N$  denotes the total number of patches. We performed maximum pooling (MAP) on each patch to obtain the feature vector.
\begin{equation}
\begin{aligned}
\bm f_{v,i}^l = {\text{MAP}}(\bm P_{v,i}^l)\quad(1 \leq l \leq N)
\end{aligned},
\end{equation}

Given the $j$-th text description $\bm X_{t,j} \in {\mathbb{R}^{M \times D}}$ in a batch, where $M$ represents the number of words and $D$ represents the dimensionality of each word vector. The input is fed into the pre-trained BERT \cite{30} model to obtain the text features $\bm T_{j} = (\bm t_{j,0}, \bm t_{j,1}, ... \bm t_{j,M}) \in \mathbb{R}^{(M+1) \times D}$, where $\bm t_{j,0}$  represents the global features (i.e., class tokens) of $\bm X_{t,j}$  and $\bm t_{j,1},..., \bm t_{j,M}$ represents the features of $M$ words. We performed max-pooling on $\bm T_{j}$  to obtain the feature  $\bm t_{j, p}$. Different sentences describing the same pedestrian image often show differences in sentence structure, making it difficult to match the features of the keyword and the corresponding local area features of the image.

\begin{figure}[t!]
\centering
\includegraphics[width=2.8in,height=2.3in]{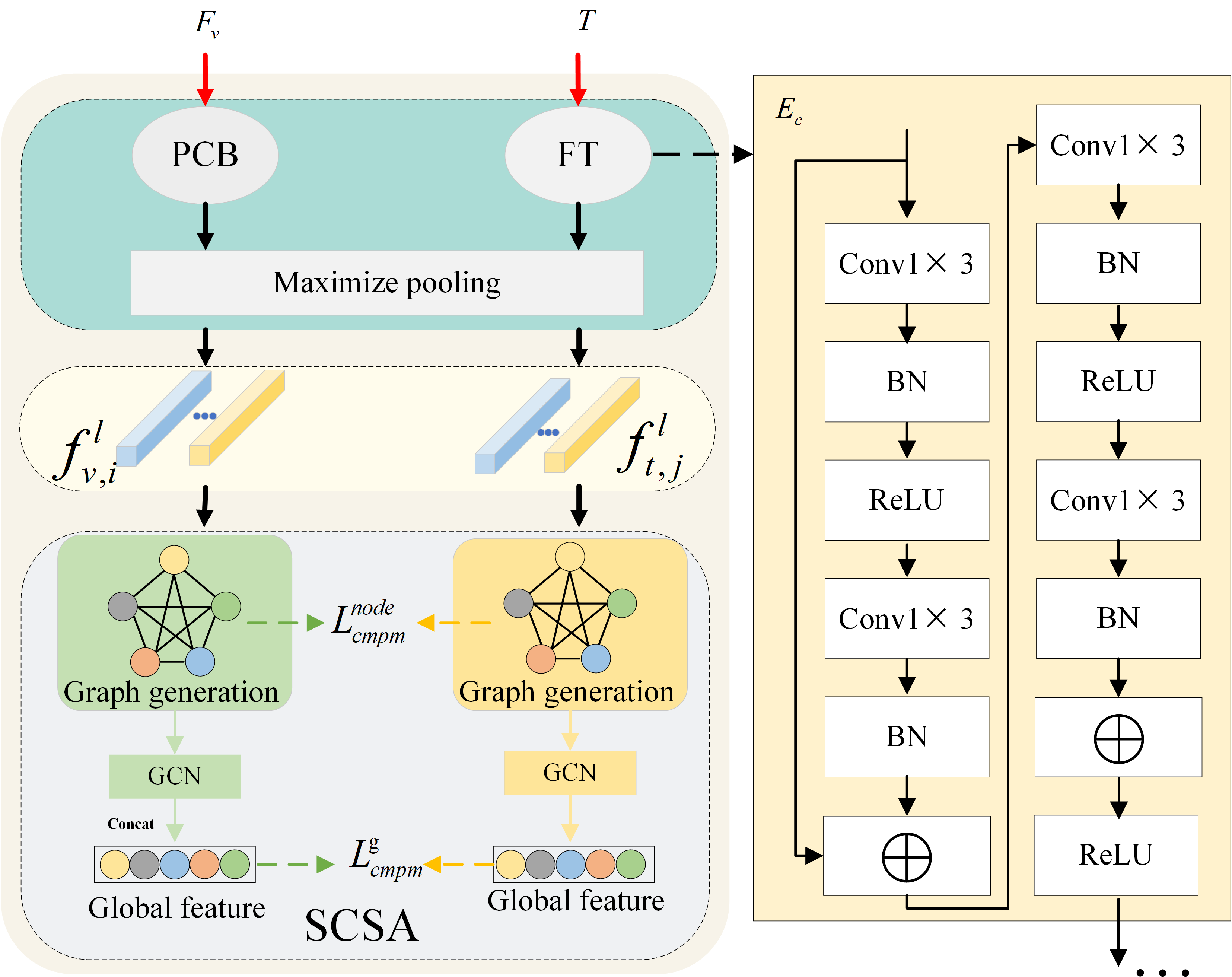}
\caption{Semantic consistency feature construction network.}
\label{fig}
\end{figure}

To address this problem, this study proposed an SCFC module. As shown in Fig. \ref{fig}, SCFC is mainly composed of a PCB, feature transformation (FT), and structure consistent semantic alignment (SCSA). PCB is used to divide pedestrian images into different patches, FT is utilized to map the global features of text to local features corresponding to specific areas of pedestrian image, and SCSA is employed to construct new global features with consistent semantic and graph structures. In this module, the feature transformation network $\bm E_c$ contains multiple FTs, and each one is employed to convert $\bm T_{j}$  to specific features that are semantically consistent with that of the specific area in a person image, thus giving text features interpretability. Each FT subnetwork in $\bm E_c$ shares the same structure but with different parameters.

The FT structure consists of four convolutional layers with $1 \times 3$ convolutional kernels, batch normalization (BN), ReLU activation function, and jump connections. The encoder $\bm E_{c}$ contains $N$ FT networks, which correspond to $N$ nodes of the pedestrian image. Let $\bm T_{j}$  denote the $j$-th text feature in a batch. After $\bm T_{j}$ is input into the  $l$-th FT, the output result is denoted as $\bm f_{t,j}^l$,where $t$ denotes the text modality. In our method, we denote the $l$-th node features of the image and text on the graph by $\bm f_{v,i}^l$ and $\bm f_{t,j}^l$, respectively. To eliminate the modality difference between $\bm f_{v,i}^l$ and $\bm f_{t,j}^l$, we introduced cross-modal projection matching (CMPM) loss \cite{33}. The feature representations of the different modalities can be associated by merging the cross-modal projections into the KL scatter. With $\bm f_{v,i}^l$ and $\bm f_{t,j}^l$, we construct the image-text pair $\{(\bm f_{v,i}^l, \bm f_{t,j}^l)\} _{l = 1}^N$. The matching probability between $\bm f_{v,i}^l$ and $\bm f_{t,j}^l$, can be calculated as
\begin{equation}
\begin{aligned}
p_{v2t,i,j}^l = \frac{{\exp ({{(\bm f_{v,i}^l)}^T}\bar{\bm f}_{t,j}^l)}}
{{\sum\nolimits_{j = 1}^n {\exp ({{(\bm f_{v,i}^l)}^T}\bar{\bm f}_{t,j}^l)}}}
\end{aligned},
\end{equation}
where $\bar{\bm f}_{t,i}^l = \frac{{\bar {\bm f}_{t,i}^l}} {{{{\left\| {\bar{\bm f}_{t,i}^l} \right\|}_2}}}$. With $p_{v2t,i,j}^l$, the CMPM loss can be formulated as:
\begin{equation}
\begin{aligned}
\ell_{v2t}^{node}({\bm E_{bv}, \bm E_{c}}) = {1 \over n}\sum\limits_{l = 1}^N {\sum\limits_{i = 1}^n {\sum\limits_{j = 1}^n {p_{v2t,i,j}^l\log ({{p_{v2t,i,j}^l} \over {q_{i,j}^l + \varepsilon }})} } }
\end{aligned},
\end{equation}
where $n$ is the batchsize, $\varepsilon=10^{-8}$ is used to avoid the zero denominator, $q_{i,j}^l = \bm y_{i,j}^l/\sum_{j = 1}^N {\bm y_{i,j}^l}$, and $\bm y_{i,j}^l = 1$ denotes that $\bm f_{v,i}^l$ and $\bm f_{t,j}^l$ have the same pedestrian identity and $\bm y_{i,j}^l = 0$ indicates that $\bm f_{v,i}^l$ and $\bm f_{t,j}^l$ have different pedestrian identities. This process narrows the distance between each visual node feature and its matching text node feature in the v2t direction. A similar process is performed through t2v to narrow down each text node feature and its corresponding visual node feature. Therefore, the bidirectional CMPM loss can be expressed as
\begin{equation}
\begin{aligned}
\ell_{cmpm}^{node}({\bm E_{bv}}, \bm E_{c}) = \ell_{v2t}^{node}({\bm E_{bv}}, \bm E_{c}) + \ell_{t2v}^{node}({\bm  E_{bv}}, \bm E_{c})
\end{aligned}.
\end{equation}

Let $\bm V_{v,i} = \{\bm f_{v,i}^1, \bm f_{v,i}^2,..., \bm f_{v,i}^N\} $ be a node set of a person image, where $\bm f_{v,i}^l$ is the $l$-th node feature. The edge set between the nodes is denoted as $\bm A_{v,i}$. Using $\bm V_{v,i}$ and $\bm A_{v,i}$, we construct a structured graph $\bm G_v= (\bm A_{v,i}, \bm V_{v,i})$. Similarly, we can construct a graph ${\bm G_{t,j}} = ({\bm A_{t,j}}, {\bm V_{t,j}})$ for the $j$-th text. The result of ${\bm V_{v,i}}$ after the $(l'+1)$-th graph convolution can be expressed as:
\begin{equation}
\begin{aligned}
\bm V_{v,i}^{l'+1}=\textrm{Sigmoid}({\bm A_{v,i}}\bm V_{v,i}^{l'}{\bm W^{l'}})
\end{aligned},
\end{equation}
where $\bm V_v^0 = \bm V_{v,i}$, $l'$ denotes the number of layers of the graph convolution, $\bm W^{l'}$ denotes the learnable matrix of the $l'$-th layer graph convolution and $\bm W^{0'}$ is the result of random initialization. We stitch together the different node features in a specific order to form semantically consistent global features $\bm f_{v,i}^g$ and $\bm f_{t,j}^g$. To ensure the discrimination of $\bm f_{v,i}^g$ and $\bm f_{t,j}^g$, the following loss function is used to optimize the network parameters in this paper:
\begin{equation}
\begin{aligned}
\ell_{cmpm}^g(\bm E_{bv}, \bm E_{c}, \bm E_{g}) = \ell_{v2t}^g(\bm E_{bv}, \bm E_{c}, \bm E_{g}) + \ell_{t2v}^g(\bm E_{bv}, \bm E_{c}, \bm E_{g})
\end{aligned},
\end{equation}
where $\bm E_{g}$ denotes the encoder composed of graph convolution, $\ell_{v2t}^g(\bm E_{bv}, \bm E_{c}, \bm E_{g})$ and $\ell_{t2v}^g(\bm E_{bv}, \bm E_{c}, \bm E_{g})$ can be similarly obtained from Eq. (3).
\subsection{Attack Node Implantation and Learning}
In this paper, we introduce an attack node to perturb the features of other nodes to provide support for the subsequent adversarial training of the matching model and make the model more robust to the diversity of text and pedestrian images.
\subsubsection{Attack Node Implantation}
We implemented adversarial perturbation for each node by introducing a single attack node. To successfully disturb the features of the other nodes, the introduced attack node should cause other nodes to be misclassified after the GCN. To obtain such an attack node, we take a randomly initialized node $\bm x^{attack}$ as the initial value of the attack node and connect it to other nodes through a new adjacency matrix $\bm A'_{m}$ for mobility $m$ ($m=v, t$).  The graph with the attack node $\bm x^{attack}$ implanted can be represented as $\bm G'_{m,i} = (\bm A'_{m,i}, \bm V'_{m,i})$, where the adjacency matrix $\bm A'_{m,i}$ for the $i$-th sample and the corresponding set of nodes $\bm V'_{m,i}$ can be represented as:
\begin{equation}
\begin{aligned}
{\bm A'_{m,i}} = \left[ {\begin{array}{*{20}{c}}
   \bm A_{m,i} & \bm I  \\
   {{\bm I^T}} & \bm 0  \\
\end{array}} \right]
\end{aligned},
\end{equation}
\begin{equation}\small
\begin{aligned}
\bm V'_{m,i} = [{\bm V_{m,i}}, \bm x^{attack}]
\end{aligned},
\end{equation}
where $\bm I \in {\mathbb{R}^{N \times 1}}$ is a column vector with all values of 1 and $\bm V_{m,i}$ is the adjacency matrix of $i$-th sample in a batch.
\begin{figure}[t!]
\centering
\includegraphics[width=3.0in,height=1.8in]{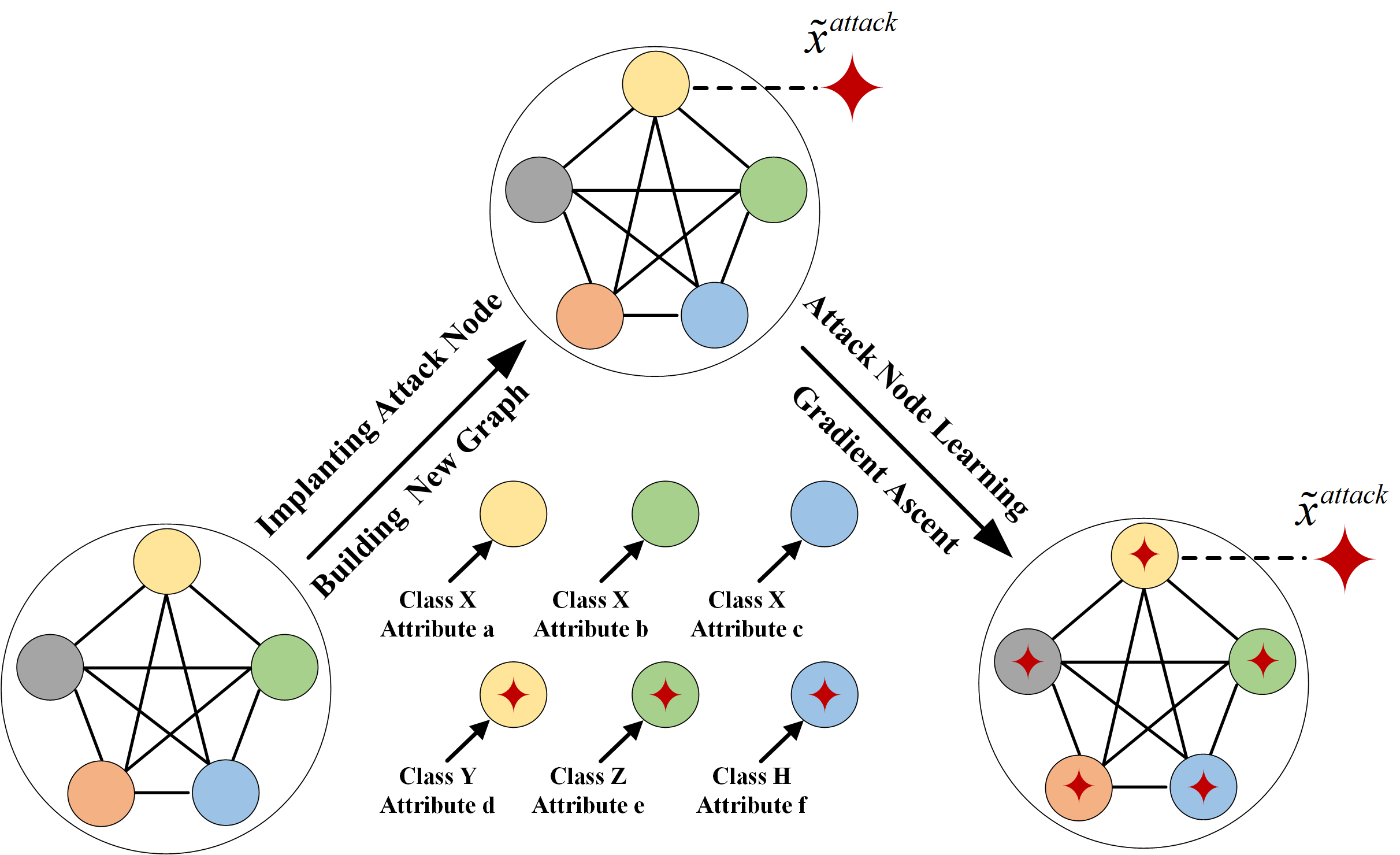}
\caption{Implanting and learning of Attack Node.}
\label{png55}
\end{figure}

\subsubsection{Attack Node Learning}
According to the above analysis, after the attack node $\bm x^{attack}$ is implanted, 
the set of node features of the $i$-th sample with mobility information $m$ generated by the $l$-th graph convolution can be expressed as $\tilde V_{m,i}$:
\begin{equation}
\begin{aligned}
{\bm V'}_{m,i}^{l}=\textrm{Sigmoid}({\bm A'}_{m,i}^{(l-1)} {\bm V'}_{m,i}^{(l-1)} {\bm W'}_{m,i}^{(l-1)}), m= v, t
\end{aligned},
\end{equation}
where $m$ denotes the modality information, $v$ and $t$ represent image and text mobilities, respectively. ${\bm V'}_{m,i}^{0}={\bm V'}_{m,i}$, ${\bm A'}_{m,i}^{0}={\bm A'}_{m,i}$, ${\bm W'}_{m,i}^{0}={\bm W'}_{m,i}$.

For the $i$-th image and the $j$-th text in one batch, the sets of node features obtained after attack node implantation and the $l$-th graph convolution can be expressed as ${\bm V'}_{v,i}^{l} = \{{\bm f'}_{v,i}^{l, 1}, {\bm f'}_{v,i}^{l, 2}, \cdots, {\bm f'}_{v,i}^{l, N}\} $ and ${\bm V'}_{t,j}^{l} = \{{\bm f'}_{t, j}^{l, 1}, {\bm f'}_{t,j}^{l, 2}, \cdots, {\bm f'}_{t,j}^{l, N}\} $.  The similarity between nodes ${\bm f'}_{v,i}^{l, k}$ and ${\bm f'}_{t, j}^{l, k}$ can be expressed as:
 \begin{equation}\small
\begin{aligned}
S_{v2t,i,j}^{l,k} = \frac{{\exp ({{({\bm f'}_{v,i}^{l,k})}^T}(\bar {\bm f'}_{t,j}^{l,k}))}}
{{\sum\nolimits_{j=1}^n {\exp ({{({\bm f'}_{v,i}^{l,k})}^T}(\bar {\bm f'}_{t,j}^{l,k}))} }}
\end{aligned},
\end{equation}
where $k$  is the ID number of the node. Text and image with the same ID number contain the same semantic information. $\bar {\bm f'}_{t,j}^{l,k}= {\bm f'}_{t,j}^{l,k}/\| {\bm f'}_{t,j}^{l,k}\|_2$. In Eq. (10), once the node sets ${\bm V'}_{v, i}^{l}$ and ${\bm V'}_{t,j}^{l}$ are injected with adversarial perturbation, the similarity $S_{v2t,i,j}^{l,k}$ between nodes ${\bm f'}_{v,i}^{l, k}$ and ${\bm f'}_{t, j}^{l, k}$ should decrease. To ensure the effectiveness of the attacked node, the CMPM loss shown in Eq. (11) is adopted to optimize the parameters of the model:
  \begin{equation}
\begin{aligned}
\ell_{cmpm}^{attack}(\bm x^{attack}_{v,i}, \bm x^{attack}_{t,j})=\ell_{v2t}^{attack}(\bm x^{attack}_{v,i}) +\ell_{t2v}^{attack}(\bm x^{attack}_{t,j})
\end{aligned},
\end{equation}
where $\ell_{v2t}^{attack}(\bm x^{attack}_{v,i})$ and $\ell_{t2v}^{attack}(\bm x^{attack}_{t,j})$ are defined respectively as:
\begin{equation}
\begin{aligned}
\ell_{v2t}^{attack}(\bm x^{attack}_{v,i}) =  - \frac{1}
{n}\sum\limits_{k = 1}^N {\sum\limits_{i = 1}^n {\sum\limits_{j = 1}^n {S_{v2t,i,j}^k\log (\frac{{S_{v2t,i,j}^k}}
{{q_{i,j}^l + \varepsilon }})} } }
\end{aligned},
\end{equation}
\begin{equation}
\begin{aligned}
\ell_{t2v}^{attack}(\bm x^{attack}_{t,j}) =  - \frac{1}
{n}\sum\limits_{k = 1}^N {\sum\limits_{j = 1}^n {\sum\limits_{i = 1}^n {S_{t2v,j,i}^k\log (\frac{{S_{t2v,j,i}^k}}
{{q_{j,i}^k + \varepsilon }})} } }
\end{aligned}.
\end{equation}
Therefore, according to the loss function in Eq. (11), we can obtain the attack node $\bm x_{attack,i}$ and $\bm x_{attack,j}$:
\begin{equation}\small
\begin{aligned}
\{\tilde{\bm x}^{attack}_{v,i}, \tilde{\bm x}^{attack}_{t,j}\}= \mathop {\arg \max }\limits_{\bm x^{attack}_{v,i}, \bm x^{attack}_{t,j}} \left\{{\ell_{v2t}^{attack}(\bm x^{attack}_{v,i})+\ell_{t2v}^{attack}(\bm x^{attack}_{t,j})} \right\}
\end{aligned}.
\end{equation}
Using the above procedure, we obtain the attack nodes $\tilde {\bm x}_{v, i}^{attack}$ and $\tilde {\bm x}_{t, j}^{attack}$ for the $i$-th image and $j$-th text in a batch, respectively.
\subsection{Adversarial Training and Algorithms}
To improve the  defense against image and text diversity, we use the learned attack node $\tilde {\bm x}_{m, i}^{attack}$ to implant adversarial perturbation information for each nodes, and train the matching model against the attacks from the adversarial perturbation. We inject trained attack nodes $\tilde {\bm x}_{v, i}^{attack}$ and $\tilde {\bm x}_{t, i}^{attack}$ on graphs ${\bm G_{v,i}} = ({\bm A_{v,i}}, {\bm V_{v,i}})$ and ${\bm G_{t,i}} = ({\bm A_{t,i}}, {\bm V_{t,i}})$ with consistent semantics and structure, respectively, which can be denoted as $\bm G_{v,i}^{attack} = (\bm A_{v,i}^{}, \bm V_{v,i}^{attack})$ and $\bm G_{t,i}^{attack} = (\bm A_{t,i}, \bm V_{t,i}^{attack})$,  where $\bm V_{v,i}^{attack}$ and $\bm V_{t,i}^{attack}$ are defined respectively as:
 \begin{equation}
\begin{aligned}
\bm V_{v,i}^{attack} = [{\bm V_{v,i}}, \tilde {\bm x}_{v,i}^{attack}], \bm V_{t,j}^{attack} = [{\bm V_{t,j}}, \tilde {\bm x}_{t,j}^{attack}]
\end{aligned}.
\end{equation}
After the $l$-th layer graph convolution, we obtain the node feature sets $\tilde{\bm V}_{v,i}^{attack, l}=\{\tilde{\bm f}_{v,i}^{l, 1}, \tilde{\bm f}_{v,i}^{l, 2}, \cdots, \tilde{\bm f}_{v,i}^{l, N}\}$ and $\tilde{\bm V}_{t,i}^{attack, l}=\{\tilde{\bm f}_{t,i}^{l, 1}, \tilde{\bm f}_{t,i}^{l, 2}, \cdots, \tilde{\bm f}_{t,i}^{l, N}\}$ injected with adversarial perturbation. We concatenate the node features $\{\tilde{\bm f}_{v,i}^{l, 1}, \tilde{\bm f}_{v,i}^{l, 2}, \cdots, \tilde{\bm f}_{v,i}^{l, N}\}$ and $\{\tilde{\bm f}_{t,j}^{l, 1}, \tilde{\bm f}_{t,j}^{l, 2}, \cdots, \tilde{\bm f}_{t,j}^{l, N}\}$ respectively, and obtain the concatenated features $\bm p_{v,i}^{attack,l}$ and $\bm p_{t,i}^{attack,l}$. Accordingly, we can obtain the similarity of $\bm p_{v,i}^{attack,l}$ and $\bm p_{t,i}^{attack,l}$:
 \begin{equation}
\begin{aligned}
p_{i,j}^{attack} = \frac{{\exp ({{(\bm p_{v,i}^{attack})}^T}\bm p_{t,j}^{attack})}}
{{\sum\nolimits_{j = 1}^n {\exp ({{(\bm p_{v,i}^{attack})}^T}\bm p_{t,j}^{attack})} }}
\end{aligned}.
\end{equation}
In adversarial training, the loss function used in this method can be expressed as:
\begin{equation}\small
\begin{aligned}
\ell^{attack}({\bm E_{bv}}, {\bm E_c}, \bm W_{attack}) &= \ell_{{cmpm}}^s{\text{(}}{\bm E_{bv}}{\text{) + }}{\lambda _1}\ell_{cmpm}^{node}({\bm E_{bv}}, {\bm E_c}) \\
&+ {\lambda _2}\ell_{cmpm}^{attack}({\bm E_{bv}},{\bm E_c}, \bm W_{attack})
\end{aligned},
\end{equation}
where $\bm W_{attack}$ is the learnable parameter in the graph convolution during adversarial training, $\ell_{{\text{cmpm}}}^s$ is the CMPM loss between $\bm t_{j, p}$ and $\bm f_{v,i}^s$ obtained by MAP of of $\bm F_{v,i}^3$, and $\ell_{cmpm}^{attack}$ is defined as:
\begin{equation}\small
\begin{aligned}
\ell_{cmpm}^{attack}(\bm E_{bv}, \bm E_{c}, \bm W_{attack})&=\frac{1}{n^{2}}\sum_{j=1}^{n}\sum_{i=1}^{n}p_{i,j}^{attack,l}\log(\frac{p_{i,j}^{attack}}{q_{i,j}+\varepsilon})\\&+p_{j,i}^{attack}\log(\frac{p_{j,i}^{attack}}{q_{j,i}+\varepsilon})
\end{aligned}.
\end{equation}
To facilitate an understanding of the proposed method, we provide the training process of the model in detail in \textbf{Algorithm} 1.
\begin{algorithm}[!t]\footnotesize
 \caption{\textbf{}Semantic Consistency Feature Construction, Attack Node Learning and Adversarial Training Algorithm}\label{alg:A}
\begin{algorithmic}
\STATE {\textbf{Input:} Image set $\bm X=\{\bm X_{i}\}_{i=1}^{n}$,  text set  $\bm T=\{\bm T_{i}\}_{i=1}^{n}$, the corresponding pedestrian identity labels  $\bm Y=\{\bm y_{i}\}_{i=1}^{n}$.\\}
\STATE {\textbf{Output:} Encoders $\bm E_{bv}$, $\bm E_{c}$ and $\bm W_{attack}$.\\
\begin{flushleft}
\textbf{Step \uppercase\expandafter{\romannumeral1}:} Semantic consistency feature construction
 (Sec.\uppercase\expandafter{\romannumeral3}.B)\\
~1:Sample a batch of  data.\\
~2:Initialize ${\bm E_{bv}}$, ${\bm E_c}$, ${\bm W^{l'}}$.\\
~3:\textbf{for} \emph{iter}=1, $\cdots$, \emph{Iteration}$_{1}$ \textbf{do}\\
~4:\qquad  Update ${\bm E_{bv}}$, ${\bm E_c}$ by minimizing the loss in Eq.(4).\\
~5:\qquad  Update ${\bm E_{bv}}$, ${\bm E_c}$ and ${\bm E_g}$ by minimizing the loss in Eq.(6) .\\
~6:\textbf{end for}\\
\textbf{Step \uppercase\expandafter{\romannumeral2}:}  Attack Node Implantation and Learning
 (Sec.\uppercase\expandafter{\romannumeral3}.C)\\
~7:Sample a batch of  data.\\
~8:Load the learned encoders ${\bm E_{bv}}$, ${\bm E_c}$ and ${\bm E_g}$.\\
~9:Initialize Attack Node $\tilde {\bm x}_{v,i}^{attack}$, $\tilde {\bm x}_{t,j}^{attack}$;\\
10: \textbf{for} \emph{iter}=1, $\cdots$, \emph{Iteration}$_{2}$ \textbf{do}\\
11:\qquad Update  $\tilde {\bm x}_{v,i}^{attack}$, $\tilde {\bm x}_{t,j}^{attack}$ via Eq.(14)\\
12: \textbf{end for}\\
\textbf{Step \uppercase\expandafter{\romannumeral3}:} Adversarial Training (Sec.\uppercase\expandafter{\romannumeral3}.D)\\
13:Sample a batch of  data.\\
14:Load the learned  ${\bm E_{bv}}$, ${\bm E_c}$, ${\bm E_{g}}$, $\tilde {\bm x}_{v,i}^{attack}$, $\tilde {\bm x}_{t,j}^{attack}$\\
15: \textbf{for} \emph{iter}=1, $\cdots$, \emph{Iteration}$_{3}$ \textbf{do}\\
16:\qquad Update $\bm E_{bv}$, $\bm E_c$ and $\bm W_{attack}$ by minimizing the loss in \\\qquad~~~~Eq.(17).\\
17: \textbf{end for}\\
\end{flushleft}}
\end{algorithmic}\label{algorithm}
\end{algorithm}

\begin{table*}[!ht]\footnotesize
\centering {\caption{Table 1 Performance comparison of different methods on the CUHK-PEDES dataset. The values of CMC (\%) for different methods are presented in this table, where the optimal results are indicated in bold.}\label{tabl133}
\renewcommand\arraystretch{1.4}
\begin{tabular}{|c|c|c|c|c|c|}
\hline
Methods&Reference&~~~~Rank-1~~~ &~~~Rank-5~~~&~~~Rank-10~~~\\
\hline
GNA-RNN \cite{1}          &CVPR'17  &19.05  &--	 &53.64    \\
GLA \cite{32}             &ECCV'18   &43.58  &66.93	 &76.26       \\
CMPM+CMPC\cite{33}        &ECCV'18 &49.27  &--	 &79.27      \\
MCCL \cite{35}            &ICASSP'19  &50.58  &--	 &79.06\\
A-GANet\cite{6}           &ACM MM'19   &53.14  &74.03	 &81.95\\
Dual-path\cite{34}        &TOMM'20    &44.4  &66.26	 &75.07\\
MIA\cite{2}               &TIP'20  &53.10  &75.00	 &82.9\\
PMA  \cite{36}            &AAAI'20  &53.81  &73.54	 &81.23\\
TIMAM\cite{37}            &ICCV'20  &54.51  &77.56	 &84.78\\
ViTAA\cite{8}             &ECCV'20  &55.97  &75.84	 &83.52\\
NAFS \cite{13}            &arXiv'21  &59.94  &79.86	 &86.7\\
DSSL \cite{29}            &ACMMM'21  &59.98  &80.41	 &87.56\\
MGEL \cite{7}             &IJCAI'21  &60.27  &80.01	 &86.74\\
SSAN \cite{9}             &arXiv'21  &61.37  &80.15	 &86.73\\
TBPS(ResNet-50) \cite{38} &arXiv'21  &61.65  &80.98  &86.78\\
NAFS with RVN\cite{13}    &arXiv'21  &61.50  &81.19  &86.78\\
SUM             \cite{39} & KBS'22   &59.22  & 80.35 &87.51\\
CLFT \cite{53}                     & TIP'22   &60.10  &79.60  &86.34\\
   \textbf{Proposed} &This paper          &\bf64.12&\bf82.76&\bf88.65\\
  \hline
\end{tabular}}
\end{table*}
\section{Experiments}
\label{sec:experiment}
\subsection{Datasets and Evaluation Protocol}
To verify the effectiveness of the proposed method, the performance of the algorithm in this paper is tested on two challenging datasets, including CUHK-PEDES \cite{1} and RSTPReid \cite{29}.

\begin{figure}[t!]
\centering
\includegraphics[width=3.5in,height=1.8in]{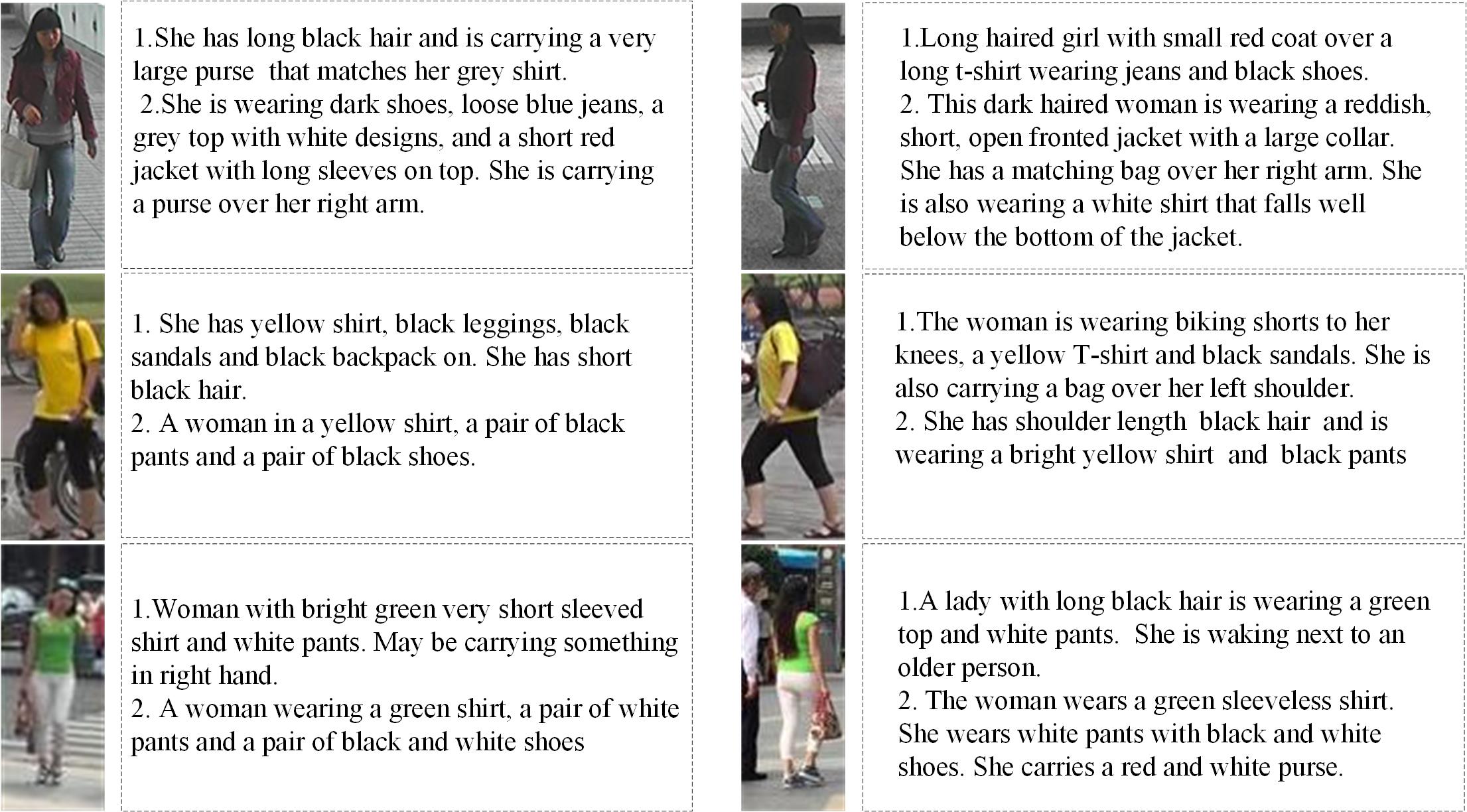}
\caption{Some samples selected from CUHK-PEDES. The left and the right text-image pairs on the same line share the same pedestrian identity. It can be seen that the images and texts of pedestrian with the same identity show diversity.}
\label{png4}
\end{figure}

\textbf{CUHK-PEDES}: This dataset is a large-scale text based pedestrian image retrieval dataset. We use the same protocol as in\cite{32} to divide the dataset into training set, validation set and test set. The training set contains 11,003 pedestrians. These pedestrians have a total of 34,054 images and 68,126 text descriptions, and some samples are shown in \ref{png4}. The validation set contains 3,078 images and 6158 text descriptions of 1,000 different identities, while the test set contains 3,074 images and 6,156 text descriptions of 1,000 different identities.

\textbf{RSTPReid}: This dataset contains 4,101 pedestrians with different identities, each with five images from different camera views, for a total of 20, 505 pedestrian images, and each image has two text descriptions. Following the dataset partitioning protocol of \cite{29}, we divide this dataset into a training set, validation set and test set. The training set contains 18, 505 images of 3, 701 pedestrians, the validation set contains 1, 000 images of 200 identity pedestrians and the test set contains 1, 000 images of 200 pedestrians. Same as the existing methods, Cumulative Match Characteristic (CMC) is used in this paper to evaluate the matching performance.
\subsection{Implementation Details}
The proposed method consists of two feature extraction  modules: image feature extraction and text feature extraction. Image feature extraction uses the pre-trained ResNet50 on ImageNet\cite{19} as the backbone. The text feature extraction module uses pre-trained BERT as the backbone. The two modules are trained for a total of 100 epochs, and the Adam optimizer \cite{30} is employed to train the model. The initial learning rate is set to 0.0001, and in epochs 0$~$10, the learning rate is adjusted according to the warm-up strategy \cite{31}. In the 51st epoch, the learning rate decays to $10{\text{\% }}$. All images are resized to $384\times 128 \times 3$, and random horizontal flipping is used for data augmentation. In the experiments, the batchsize is set to 64, and each batch contains 64 image-text pairs. In the testing phase, cosine distance is used to measure the similarity of the image-text pairs. The hyper-parameters in Eq.(17) are set to ${\lambda _1} = 1$ and ${\lambda _2} = 1$, respectively. The model in this study is implemented on the PyTorch platform, and all experiments are performed on a single NVIDIA GeForce RTX3090 GPU.
\subsection{Comparison With State-of-the-Art Methods}
\textbf{Results on the CUHK-PEDES dataset}: To verify the effectiveness of the proposed method, we test our method on the CUHK-PEDES dataset and compare its performance with some state-of-the-art methods. The methods involved in the comparison include the GNA-RNN \cite{1}, GLA \cite{32}, CMPM+CMPC\cite{33}, MCCL \cite{35}, A-GANet\cite{6}, Dual-path \cite{34}, MIA\cite{2}, PMA  \cite{36}, TIMAM \cite{37}, ViTAA \cite{8}, NAFS \cite{13}, DSSL \cite{29}, MGEL \cite{7}, SSAN \cite{9}, TBPS(ResNet-50) \cite{38}, SUM \cite{39}, NAFS with RVN\cite{13} and CLFT \cite{53}  
The experimental results of the different methods are listed in Table \ref{tabl133}. The accuracies obtained by our method on Rank-1, Rank-5 and Rank-10 are 64.12\%, 82.76\% and 88.65\%, respectively, which are better than the performance achieved by all the compared methods. In addition, we found that the latest methods, MGEL, SSAN, TBPS, NAFS and CLFT, all achieve a performance of more than 60\% on Rank-1. This is owing to the use of the attention mechanism, which allows the network to be more adaptive in extracting the desired discriminative features. However, they do not consider the impact of the diversity of texts and person images on text-image matching, so the performance is somewhat limited. Compared with CLFT, which achieves the second-best performance, the performance of the proposed method can still surpass that achieved by CLFT without using a complex module similar to attention. This demonstrates the effectiveness of the proposed method and its superiority over the other methods.
\begin{table}[!ht]\footnotesize
\centering {\caption{The experimental results of the proposed method  are compared with the state-of-the-art method on RSTPReid dataset.  The values of CMC (\%) for different methods are presented in this table, and the best result is indicated in bold.}\label{tabl144}
\renewcommand\arraystretch{1.4}
\begin{tabular}{|c|c|c|c|c|c|}
\hline
   Methods             &Reference  &Rank-1 &Rank-5 &Rank-10\\
\hline
IMG-Net\cite{42}       &JEI'20     &37.60   &61.15  &73.55\\
AMEN\cite{43}          &PRCV'21    &38.45   &62.40  &73.80\\
DSS\cite{29}           &ACMMM'21   &39.05   &62.60  &73.95\\
SSAN\cite{9}           &arXiv'21   &43.50   &67.80  &77.15\\
SUM\cite{39}           &KBS'22     &41.38   &67.48  &76.48\\
\textbf{Proposed} &This paper &\bf45.88&\bf70.45&\bf81.30\\
\hline
\end{tabular}}
\end{table}

\begin{table}[!ht]\footnotesize
\centering {\caption{Ablation experiments of the proposed method.  The matching accuracies (\%) of CMC on Rank-1, Rank-5 and Rank-10 are given in the table.}\label{tabl155}
\renewcommand\arraystretch{1.4}
\begin{tabular}{|c|c|c|c|c|c|}
\hline
Methods&Rank-1 &Rank-5&Rank-10\\
\hline
   Baseline              &55.14	&76.64	 &84.48 \\
 Baseline+SCFC            &62.07	&81.47	 &87.20 	\\
 Baseline+SCFC+ANL         &57.91	&78.72	 &85.79 \\
 Baseline+ SCFC+ANL+AT     &64.12	&82.76	 &88.65 	\\
  \hline
\end{tabular}}
\end{table}

\textbf{Results on the RSTPReid dataset}: Because the RSTPReid dataset is only recently published, there are fewer comparable methods on this dataset.  In this experiment, we use only the latest five methods, that are IMG-Net\cite{42}, AMEN\cite{43}, DSS\cite{29}, SSAN\cite{9} and SUM\cite{39}, 
to compare the performance with the proposed method. As presented in Table \ref{tabl144}, the latest method SUM 
achieves the suboptimal performance, and the recognition accuracy on Rank-1, Rank-5 and Rank-10 reaches 41.38\%, 67.48\% and 76.48\%, respectively. In contrast, the recognition performance of the proposed method is 45.55\%, 70\% and 80\%, which exceeds the performance obtained by SUM. The above experiments further demonstrate the effectiveness of the developed method.
\subsection{Ablation Study}
The method proposed in this paper mainly consists of semantic consistency feature construction (SCFC), attack node learning (ANL), and adversarial training (AT). In this study, the pre-trained BERT combined with ResNet50 obtained under the constraint of CMPM loss is used as the baseline. To verify the effectiveness of each component, different modules are added to Baseline item by item to observe changes in matching performance. In this process, we name the model obtained by adding SCFC to baseline as ``Baseline+SCFC'', and the model obtained by introducing attack nodes into ``Baseline+SCFC'' as ``Baseline+SCFC+ANL''. The complete model, containing SCFC, ANL and AT, is named ``Baseline+SCFC+ANL+AT''. All the experiments are conducted on the CUHK-PEDES dataset, and the experimental results are listed in Table \ref{tabl155}.

\textbf{Effectiveness of SCFC}: In this study, SCFC is primarily used to solve the problem of feature mismatch between text features and the corresponding image local features. As presented in Table \ref{tabl155}, the rank-1 of ``Baseline'' can only reaches 55.14\% when SCFC is not used, and when SCFC is introduced into ``Baseline'', the recognition accuracy of ``Baseline+SCFC'' on Rank-1 increases from 55.14\% to 59.32\%, which is an improvement of 4.18\%. This is mainly because SCFC can effectively eliminate the ambiguity of features caused by cross modal misalignment.

\textbf{Effectiveness of ANL}: To improve the robustness of the model to text and image diversity, a feature extraction method with the implantation of attack node is proposed. In this process, attack node implantation is mainly used to degrade the model performance to improve the  model's defense against adversarial perturbations in adversarial training. The effectiveness of the model ``Baseline+SCFC+ANL'' after adding ANL to ``Baseline+SCFC'' is given in Table \ref{tabl155}. One can that the model performance appears to be significantly reduced, which indicates that the learning method of attack node proposed in this paper is effective.

\textbf{Effectiveness of AT}: AT is mainly used to make the model more defensive against the perturbations that appear in the features. From the results in Table \ref{tabl155}, we can  observe that the matching performance of ``Baseline+ SCFC+ANL+ AT'' on Rank-1 is improved from 55.14 \% to 64.12\% after the learned attack node is injected into the graph convolution in the adversarial training. This proves the effectiveness of the proposed adversarial training in this paper.

\begin{figure*}[t!]
\centering
\includegraphics[width=6.7in,height=3.2in]{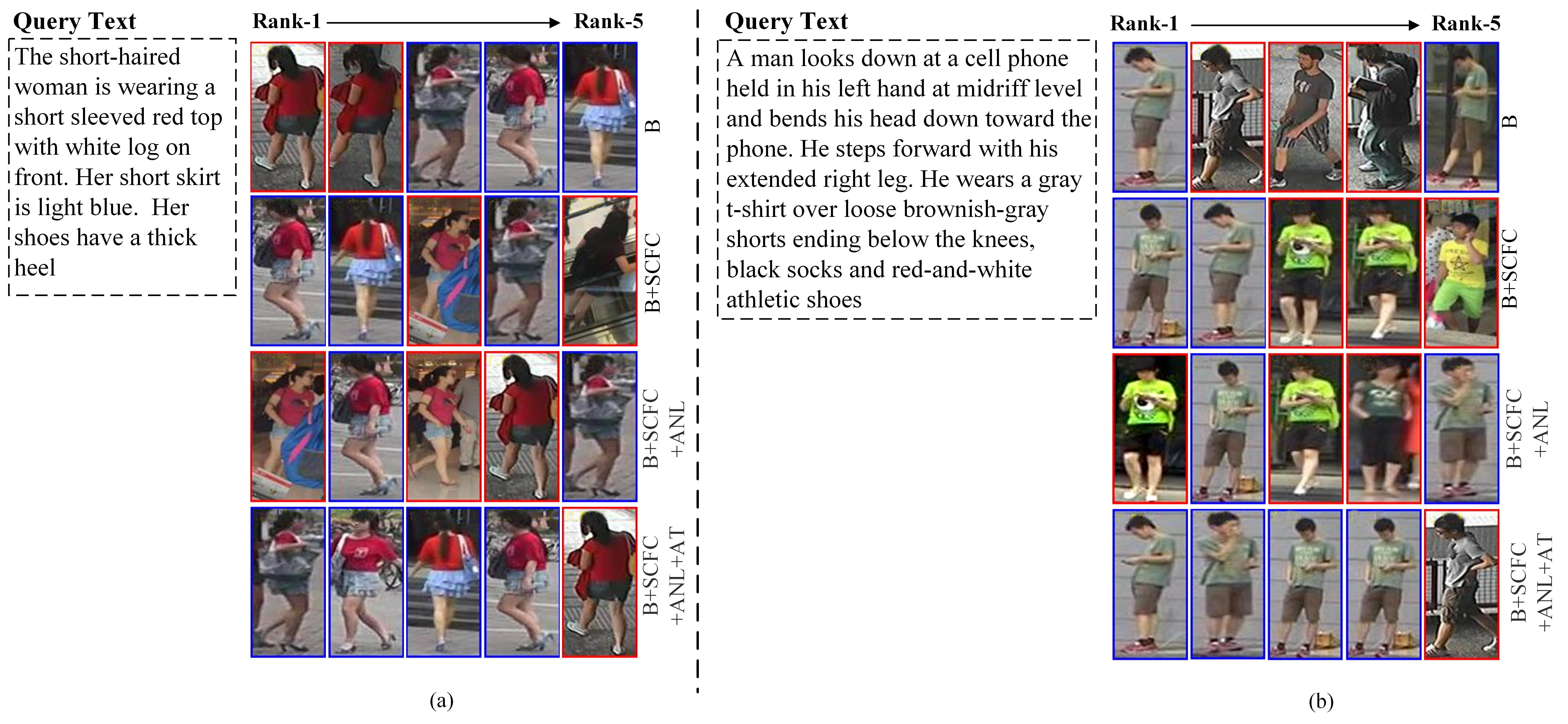}
\caption{Visualization of the effectiveness of different components. A red box indicates that the retrieval result is incorrect, a blue box indicates that the retrieval result is correct. ``B'' denotes the ``Baseline''.}
\label{png5}
\end{figure*}

\textbf{Visualization of the ablation experiment}: Fig. \ref{png5} shows the effectiveness of each component. It can be seen in Fig. \ref{png5} that the matching accuracy is improved when SCFC and AT are added to ``Baseline'' and ``Baseline+SCFC+ANL'', respectively, which proves the effectiveness of the proposed SCFC and AT. The performance of the model decreases when ANL is added to ``Baseline+SCFC'', which demonstrates the effectiveness of the proposed attack node learning. The above conclusions are consistent with those obtained from Table \ref{tabl155}.

\subsection{Further Discussion}
\begin{figure}[t!]
\centering
\subfigure[]{\includegraphics[height=1.3in,width=1.5in]{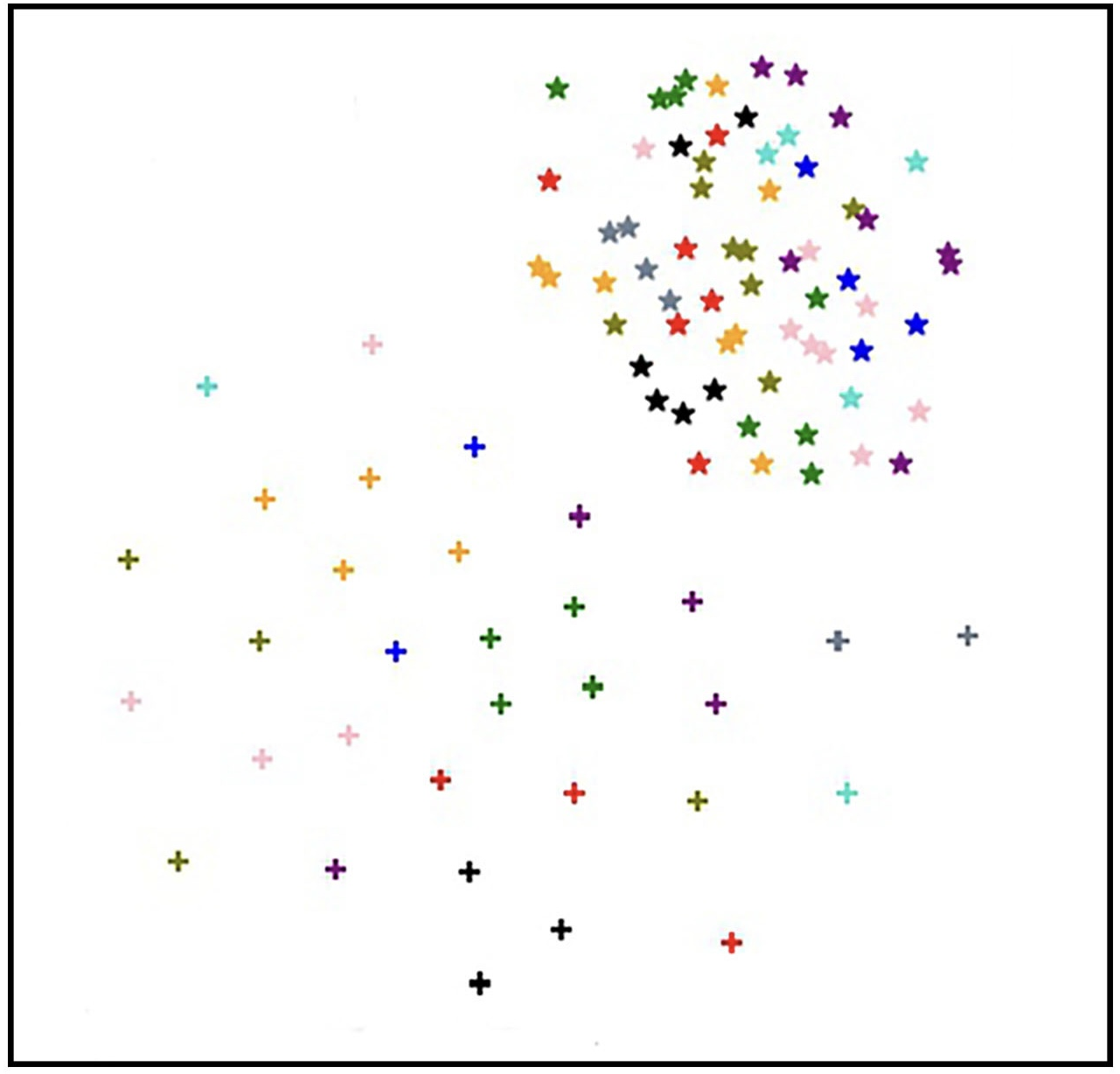}}
\subfigure[]{\includegraphics[height=1.3in,width=1.5in]{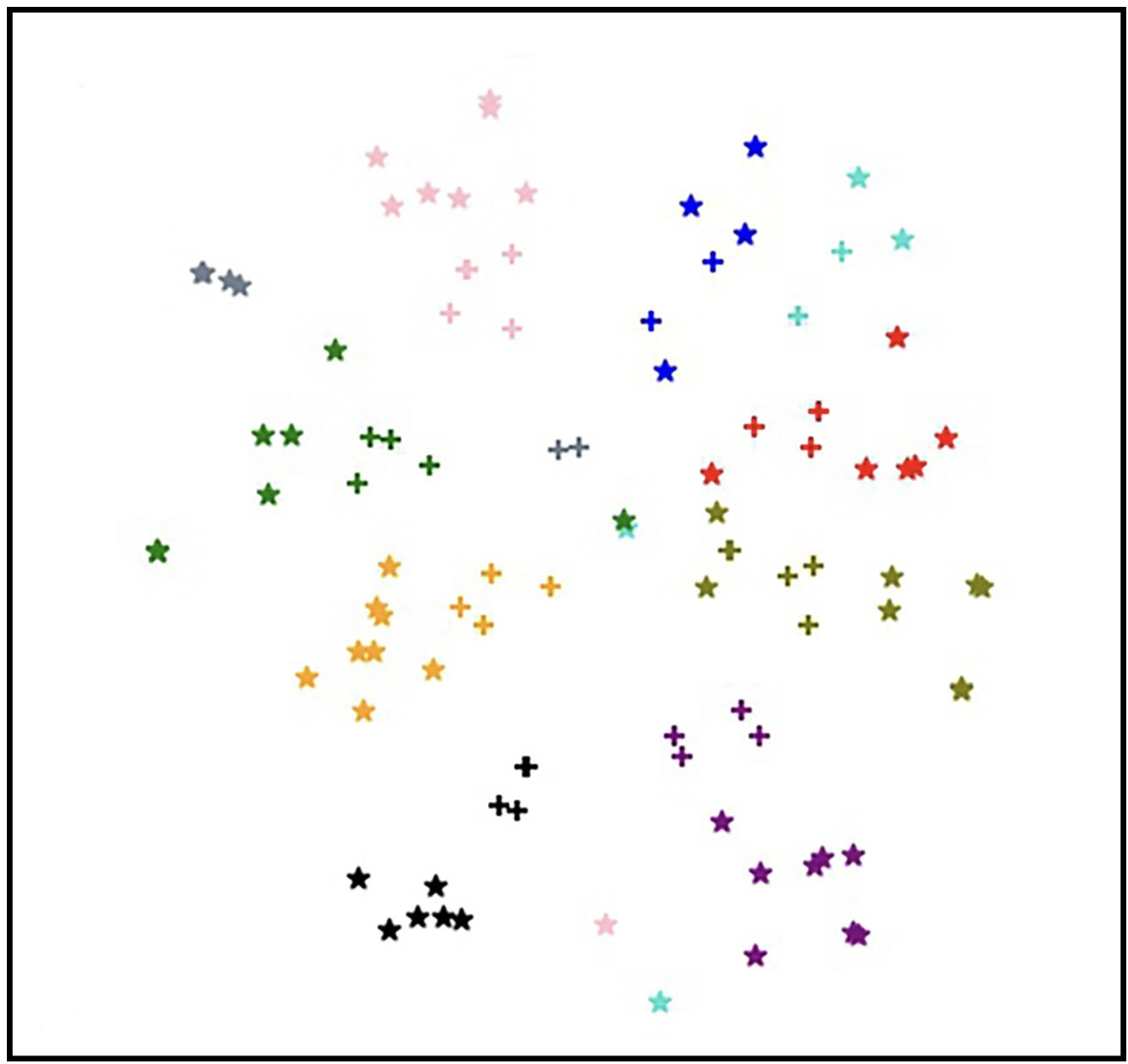}}
\caption{Visualization of model pre-training and post-training features t-SNE. (a) feature representation before model training with the proposed method, (b) feature representation after model training with the proposed method. ``+'' indicates pedestrian images and ``*'' denotes text features. The same color indicates having the same pedestrian identity.}
\label{png8}
\end{figure}

Modality discrepancy between text and images is the main challenge in matching text and pedestrian images. The proposed method can effectively reduce this discrepancy. To demonstrate the performance of the method in eliminating modal discrepancies, we use t-SNE \cite{40} to visualize the features. Fig. \ref{png8}(a) shows the features obtained without training the model using the method proposed in this study. The significant modality differences between the text and pedestrian images result in a lack of consistency between the features of text-person image pairs. As shown in Fig. \ref{png8}(b), the features obtained after training by the method in this study show greater separability for different pedestrian identities, while the features with different modalities of the same identity also show some aggregation, which is beneficial to the matching of text and pedestrian images. This indicates that the proposed method can learn feature representations with strong discrimination from person images and text. Therefore, the proposed method  can effectively reduce the challenge of modal differences in text-person image matching.

\begin{figure}[!t]
\centering
\subfigure[] {\includegraphics[height=1.2in,width=1.6in,angle=0]{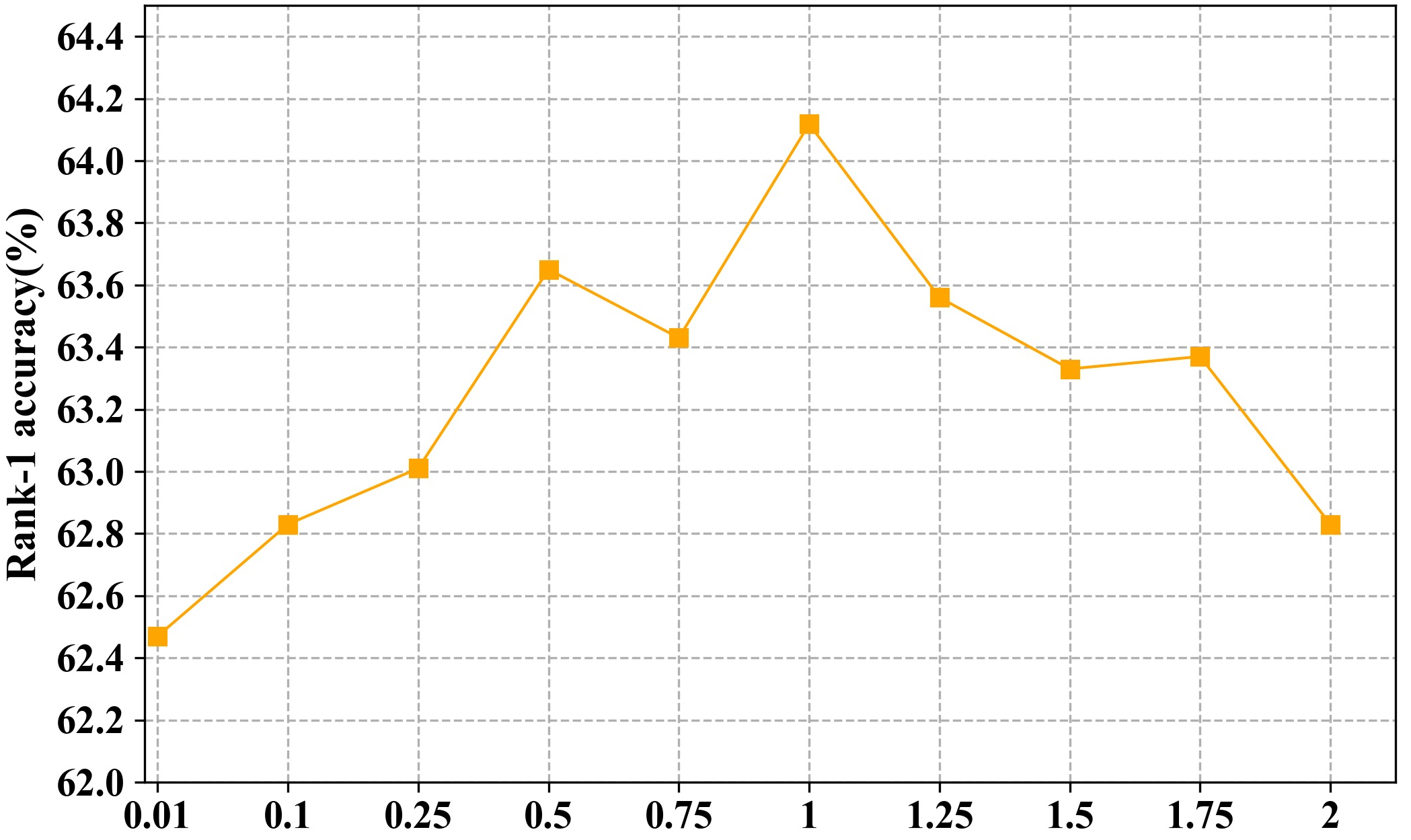}}
\subfigure[] {\includegraphics[height=1.2in,width=1.6in,angle=0]{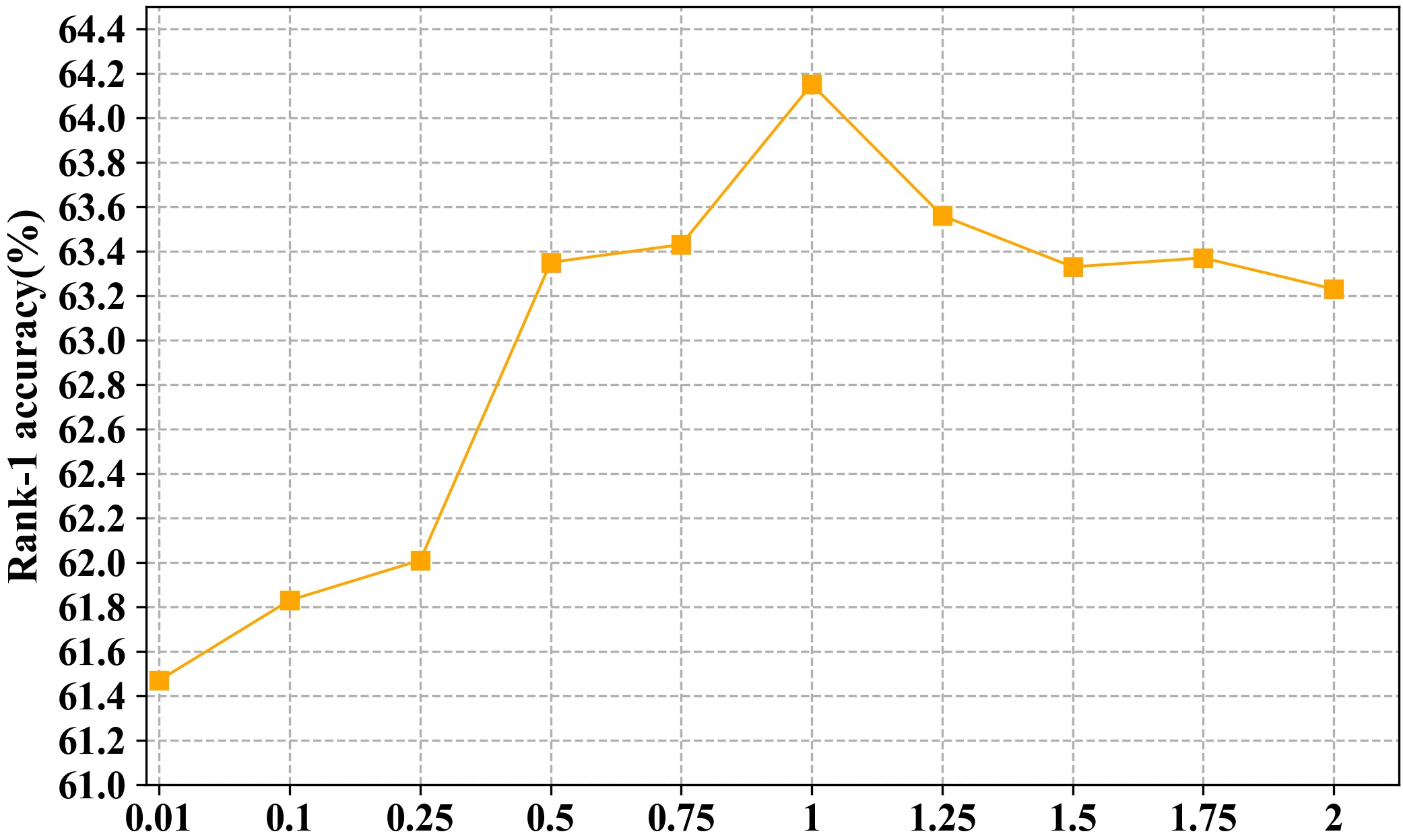}}
 \caption{Effect of ${\lambda _1}$ and ${\lambda _2}$ on model performance at different values. (a) the effect of ${\lambda _1}$ at different values; (b) the effect of ${\lambda _2}$ at different values.}
\label{png9}
\end{figure}

 \begin{figure*}[!t]
\centering
\subfigure[] {\includegraphics[height=1.8in,width=2.0in,angle=0]{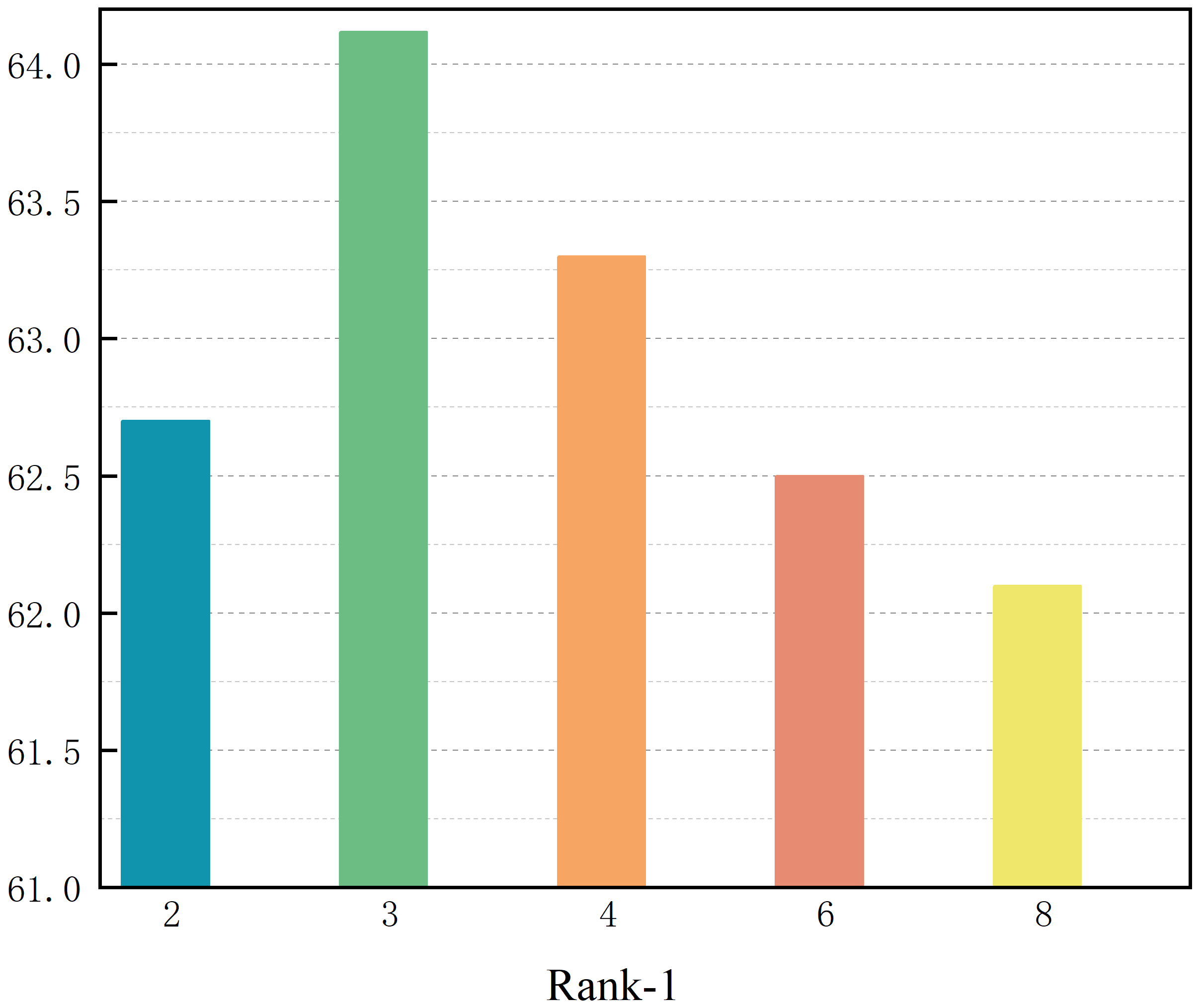}}
\subfigure[] {\includegraphics[height=1.8in,width=2.0in,angle=0]{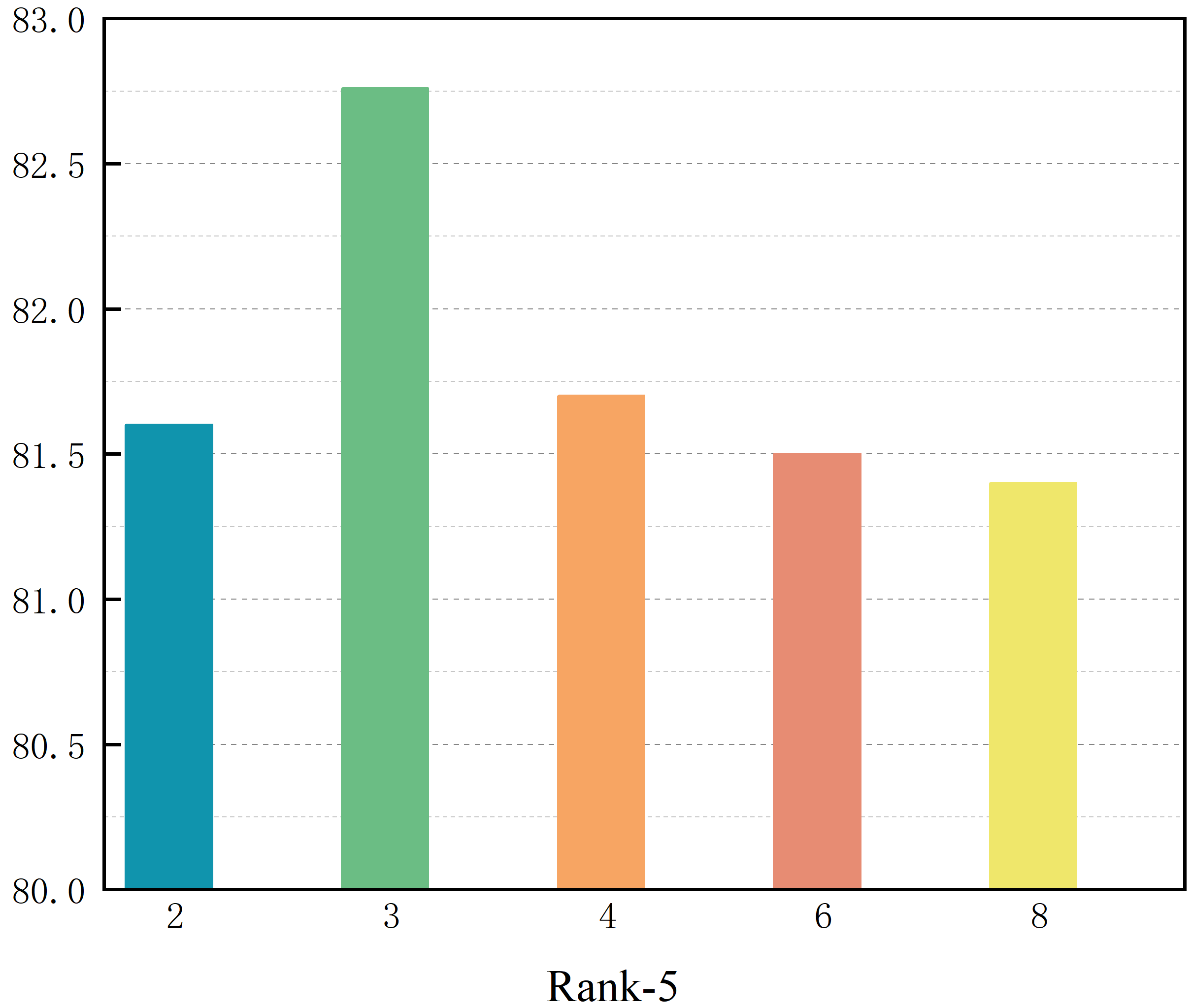}}
\subfigure[] {\includegraphics[height=1.8in,width=2.0in,angle=0]{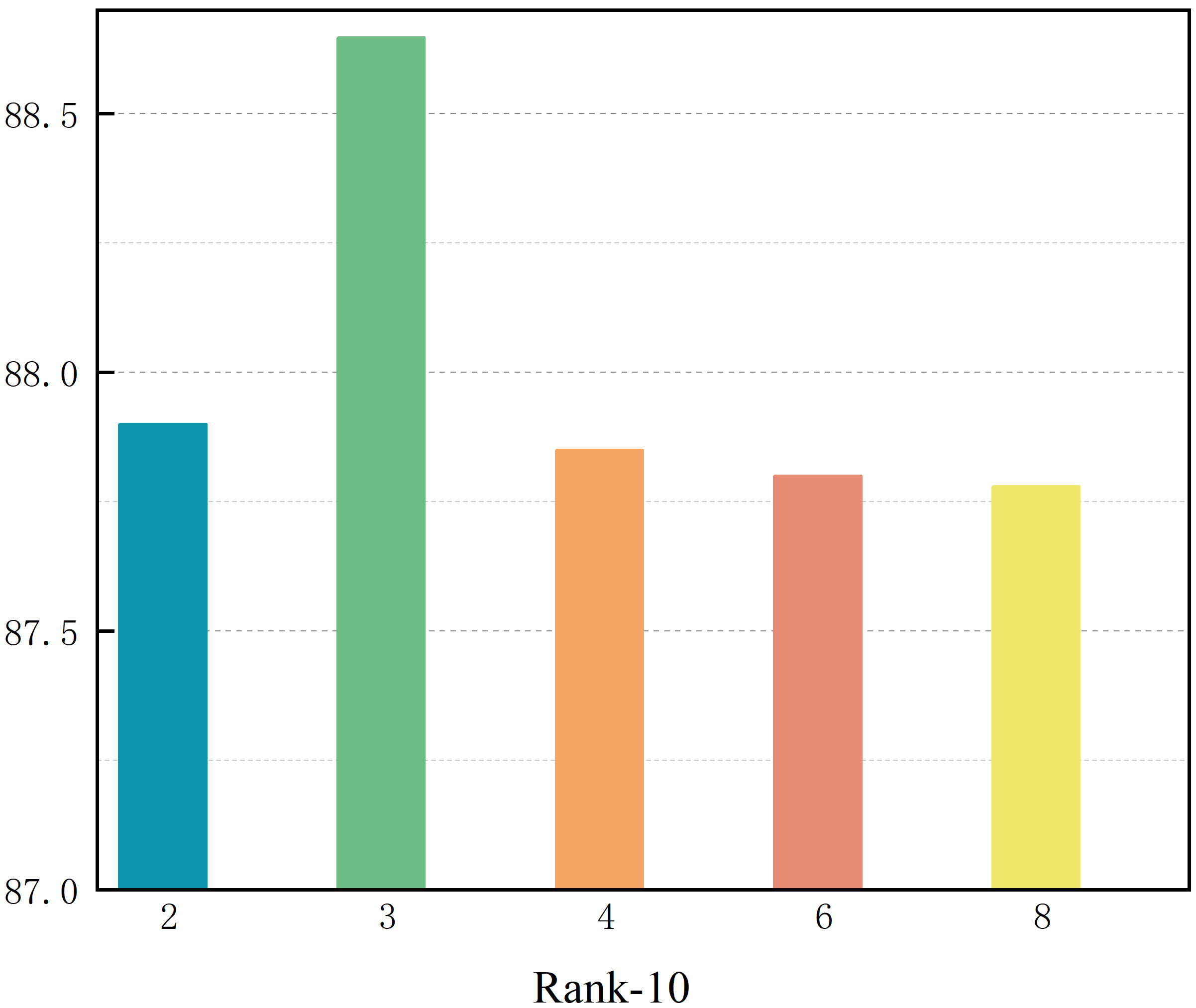}}
 \caption{Effect of $N$  on model performance when it takes different values.}
\label{png10}
\end{figure*}
\subsection{Parameter Selection and Analysis}
The approach in this paper mainly involves the hyperparameters ${\lambda _1}$, ${\lambda _2}$ in the loss function and the number of blocks $N$ in which the image features are divided. In the analysis of the role of hyperparameters, we fix two parameters to analyze the effect of another parameter on the experimental performance. In this process, all experiments are performed on the CUHK-PEDES dataset.

\textbf{The influence of} ${\lambda _1}$. Fig. \ref{png9}(a) shows the changes of the matching performance of the proposed method on CUHK-PEDES when ${\lambda _1}$ takes different values. As illustrated in Fig. \ref{png9}(a), when ${\lambda _1} \in [0.01,1]$, the matching accuracy of the proposed method on Rank-1 improves from 62.5\% to 64.12\%. When ${\lambda _1} \in [1,2]$, the matching accuracy on Rank-1 decreases from 64.12\% to 62.8\%. This indicates that ${\lambda _1}{\text{ = }}1$ is the optimal choice in our work.

\textbf{The influence of} ${\lambda _2}$ To investigate the effect of ${\lambda _2}$ on the matching performance when it takes different values, we fix  ${\lambda _1}=1$
and make ${\lambda _2}$ changes in [0, 2]. As indicated in \ref{png9} (b), the matching accuracy of the proposed method on Rank-1 gradually improves when the value of ${\lambda _2}$ is varied from 0.01 to 1, and reaches the peak when ${\lambda _2}=1$, and when ${\lambda _2}$ changes from 1 to 2, the matching performance decreases from 64.12\% to 63.2\%. It demonstrates that ${\lambda _2}=1$ is a good choice for the proposed method.

 \textbf{The influence of} $N$. In SCFC, we divide the text and image features into $N$ local features and perform semantic alignment for them. To investigate the effect of different values of $N$, we set $N$ to 2, 3, 4, 6, and 8 to analyze the variation of model performance. Fig. \ref{png10} shows the effect of $N$ choosing different vaues on the model performance. It can be seen in Fig. \ref{png10} that the model performance  is optimal when $N$ is set to 3. For this reason, we set the value of $N$ to 3 throughout this paper.

\section{Conclusion}
\label{sec:conclusion}
In this study, we propose a new framework for text-person image matching. First, we propose a semantic consistency feature construction method to achieve the semantic alignment of cross modal features, which solves the problem of inconsistent semantic information between text and person image features. To further improve the feature representation, we investigate how to reduce the challenge of text and pedestrian image diversity for their matching, which is significant for larger scale text-based person image retrieval but remains under-explored. To solve this, we propose a novel solution, wherein we treat the perturbation caused by the diversity of text and person images as adversarial perturbations and propose the implantation of  adversarial attack node in graph convolution to inject perturbation information for each node. Finally, the robustness of the matching model can be improved by the perturbation via adversarial training. This empowers the model to defend against perturbations caused by  sample diversity. The experimental results  on the CUHK-PEDES and RSTPReid datasets validate the effectiveness of the proposed method. The contribution of each module is investigated through ablation studies. The results show that each component of the proposed method is suitable for text-person image matching.
\bibliography{mybibfile(Minor1)}
\end{document}